\documentclass[10pt,twocolumn,letterpaper]{article}

\usepackage{cvpr}              

\definecolor{cvprblue}{rgb}{0.21,0.49,0.74}
\usepackage[draft]{hyperref}

\def\paperID{15088} 
\def\confName{CVPR}
\def\confYear{2025}

\title{RefPose: Leveraging Reference Geometric Correspondences for Accurate 6D Pose Estimation of Unseen Objects}

\author{Jaeguk Kim$^{1}$ \quad Jaewoo Park$^{1}$ \quad Keuntek Lee$^{1}$ \quad Nam Ik Cho$^{1,2}$\\
$^{1}$Department of ECE, INMC, Seoul National University, Korea\\
$^{2}$IPAI, Seoul National University, Korea\\
{\tt\small \{jaeguk, bjw0611, leekt000, nicho\}@snu.ac.kr}
}
\begin{document}
\maketitle

\begin{abstract}
Estimating the 6D pose of unseen objects from monocular RGB images remains a challenging problem, especially due to the lack of prior object-specific knowledge. To tackle this issue, we propose RefPose, an innovative approach to object pose estimation that leverages a reference image and geometric correspondence as guidance. RefPose first predicts an initial pose by using object templates to render the reference image and establish the geometric correspondence needed for the refinement stage. During the refinement stage, RefPose estimates the geometric correspondence of the query based on the generated references and iteratively refines the pose through a render-and-compare approach. To enhance this estimation, we introduce a correlation volume-guided attention mechanism that effectively captures correlations between the query and reference images. Unlike traditional methods that depend on pre-defined object models, RefPose dynamically adapts to new object shapes by leveraging a reference image and geometric correspondence. This results in robust performance across previously unseen objects. Extensive evaluation on the BOP benchmark datasets shows that RefPose achieves state-of-the-art results while maintaining a competitive runtime.
\end{abstract}

\section{Introduction}
\label{sec:intro}

6D pose estimation is a key aspect of computer vision and robotics, focusing on accurately predicting an object's position (3D translation) and orientation (3D rotation) in a given scene. This task is essential for a range of applications, such as autonomous driving~\cite{chen2017multi, manhardt2019roi}, augmented reality (AR)~\cite{marchand2015pose, rambach20186dof}, and robotic manipulation~\cite{busam2015stereo, perez2016robot}. Despite significant research efforts in this field, estimating the 6D pose of previously unseen objects remains a considerable challenge. This challenge largely stems from the lack of prior knowledge and the limited generalization capabilities of existing models when faced with new objects~\cite{hodan2024bop}.

In instance-level object pose estimation, various methods rely on geometric correspondence as a crucial element for achieving accurate pose estimation~\cite{park2019pix2pose, cai2020reconstruct, di2021so, li2019cdpn, su2022zebrapose, zakharov2019dpod, hodan2020epos, wang2021gdr, park2024leveraging}. Geometric correspondence refers to identifying the 3D model points that correspond to each pixel in a 2D image, which provides essential information for determining an object's pose. Typically, this correspondence is estimated using deep learning networks and is subsequently used to infer the object's pose through methods such as PnP/RANSAC~\cite{lepetit2009ep} or neural network regression. However, these techniques face challenges in accurately estimating 2D-3D geometric correspondences for unseen objects, mainly due to their reliance on pre-defined object models and the limited generalization capabilities of the correspondence estimation network.

\begin{figure}[t]
  \centering
   \fbox{\includegraphics[width=0.97\linewidth]{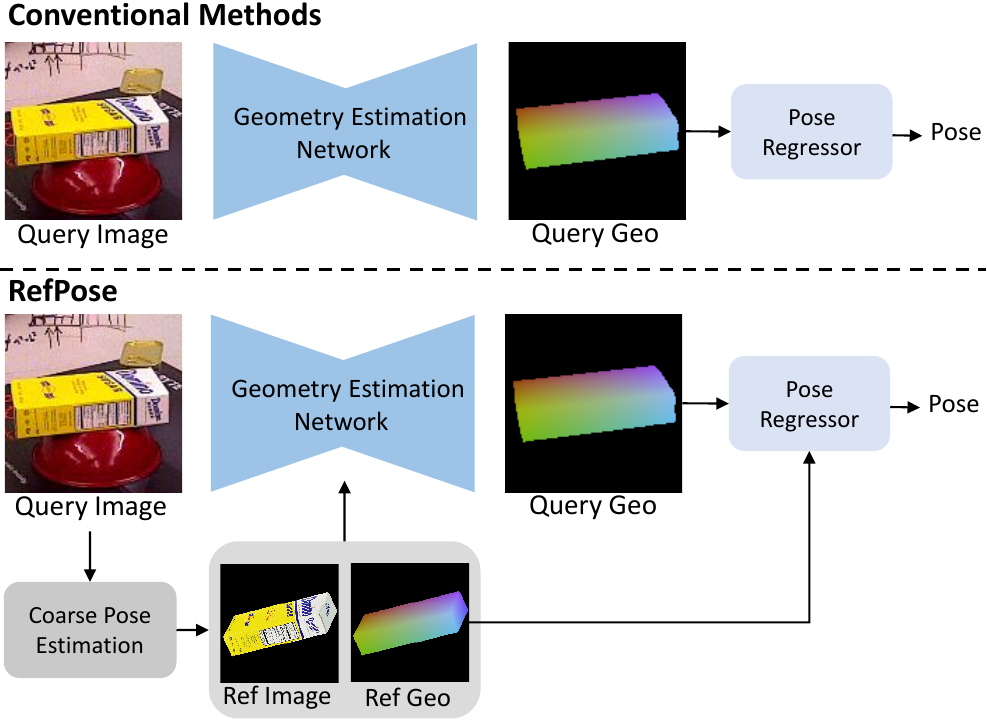}}
   \caption{Comparison between conventional methods and proposed method (RefPose). In contrast to conventional methods, RefPose leverages a reference image and geometric correspondence generated from the estimated pose in the coarse pose estimation stage to guide the query's geometric correspondence and pose estimation.}
   \label{fig:fig1}
\end{figure}

To address these limitations, we propose RefPose, a novel approach to object pose estimation for unseen objects. As illustrated in~\cref{fig:fig1}, RefPose predicts the geometric correspondence of the query object by leveraging a reference image and geometric correspondence as guidance. This guidance provides crucial geometric information about the target object, allowing the network to avoid reliance on shape priors learned from a fixed set of objects during training. 
FoundPose~\cite{ornek2025foundpose} also estimates correspondences using pre-rendered templates through patch-wise matching, similar to our approach. However, these pre-rendered templates often lack proper alignment with the query image, leading to inaccurate matches and insufficient geometric information. To overcome this, we perform coarse pose estimation by processing the pre-rendered templates to obtain an initial pose, which is then used to render a reference image closely aligned with the query image, providing more reliable information. Additionally, rather than relying solely on direct matching, we carefully design a network that enhances correspondence estimation by integrating information from both the reference image and geometric correspondence guidance.

Specifically, in the coarse pose estimation stage, we select multiple templates from a set of pre-rendered templates for the target object. This selection is based on the accuracy of optical flow predictions made by the optical flow network~\cite{teed2020raft}, which will later help us estimate the geometric correspondence for the query object. We then use these selected templates in a warping-based approach that employs medoid voting to enhance robustness against outliers, yielding a reliable coarse geometric correspondence for the query object. Subsequently, we obtain an initial pose using PnP/RANSAC and render references accordingly. With synthesized reference guidance, we estimate a more reliable and precise geometric correspondence for the query during the refinement stage. To improve this estimation, we introduce a novel attention mechanism that leverages a correlation volume from the optical flow network, effectively integrating reference information. The estimated geometric correspondence for the query then serves as a fixed basis for further refining the pose. We iteratively update the initial pose by estimating the relative pose in comparison to the reference geometric correspondence. At each iteration, we re-render the reference geometric correspondence using the updated pose and compare it to the fixed geometric correspondence of the query. This render-and-compare process is repeated until we achieve an accurate final pose.

We assess our method on seven key datasets from the BOP benchmark~\cite{hodan2024bop}. Our findings indicate that RefPose achieves superior accuracy in both coarse pose estimation and final pose accuracy compared to state-of-the-art methods. By optimizing runtime in the pose refinement stage, RefPose not only delivers the highest accuracy across all methods but also maintains competitive speed.

Our contributions are as follows:
\begin{itemize}
\item We propose RefPose, a method that leverages a reference image and geometric correspondence to guide the estimation of the query's geometric correspondence and pose, eliminating the need for shape priors learned from predefined object sets.
\item We present a classifier based on optical flow for improved template selection. Additionally, we introduce a warping-based geometry estimation method that utilizes medoid voting to enhance robustness against outliers, leading to more accurate coarse pose estimates.
\item We propose a correlation volume-guided attention mechanism, improving the model's ability to focus on relevant regions in a reference image corresponding to the query image.
\item We achieve state-of-the-art results on the BOP benchmark datasets while maintaining competitive runtime.
\end{itemize}

\begin{figure*}[t]
  \centering
   \includegraphics[width=\linewidth]{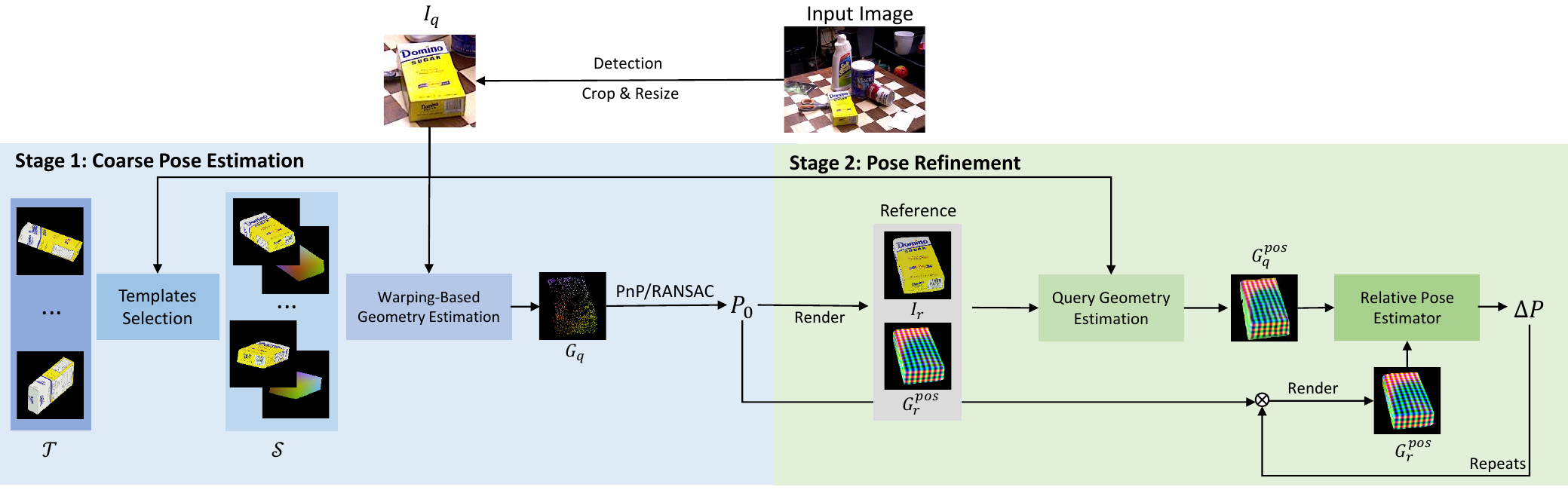}
   \caption{Overview of the RefPose pipeline. Given an input RGB image, the target object is first detected, cropped, and resized to create the query image, $I_{q}$. In \textbf{Stage 1: Coarse Pose Estimation}, a set of templates, $\mathcal{S}$, is selected from the pre-rendered template set, $\mathcal{T}$, to estimate an initial pose, $P_{0}$, for the query object. In \textbf{Stage 2: Pose Refinement}, the query's geometric correspondence, $G_{q}^{pos}$, is estimated using the rendered reference image, $I_{r}$, and geometric correspondence, $G_{r}^{pos}$. The initial pose, $P_{0}$, is iteratively refined by estimating the relative pose, $\Delta P$, between the query and reference. At each iteration, $G_{r}^{pos}$ is re-rendered to align with the updated pose, leading to an accurate final pose estimate.}

   \label{fig:fig2}
\end{figure*}

\section{Related work}
\label{sec:relatedwork}
\noindent{\textbf{Geometric correspondence-based pose estimation.}} In instance-level object pose estimation, where training and testing are performed within a fixed set of objects, a common strategy is to utilize 2D-3D geometric correspondence. Most methods follow a two-step process: they first establish 2D-3D correspondences from an RGB image and then determine the pose using a RANSAC-based PnP algorithm or a neural network. Early studies~\cite{rad2017bb8, tekin2018real} employed the 3D bounding box corners of the object as keypoints, focusing on identifying the projected positions of these points in the image. For more robust and precise pose estimation, recent research~\cite{park2019pix2pose, cai2020reconstruct, di2021so, li2019cdpn, zakharov2019dpod, hodan2020epos, wang2021gdr} has shifted toward establishing dense correspondence maps rather than relying on sparse points. Consequently, designing and training deep learning networks that can accurately predict geometric correspondence maps is essential. However, these networks often struggle with generalization, especially when applied to unseen objects outside the fixed training set. Some models~\cite{su2022zebrapose, lian2023checkerpose} even require separate network models for each object. Therefore, to effectively estimate the geometric correspondence of the query object, we train the network using not only the query image but also a reference image and its geometric correspondence, providing reliable and valuable contextual information.   

Most methods use 3D coordinates as the geometric correspondence; however, some approaches modify these coordinates to achieve finer correspondence estimation. For example,~\cite{su2022zebrapose, lian2023checkerpose} adopts a binary code representation, while~\cite{park2024leveraging} applies positional encoding, as seen in NeRF~\cite{mildenhall2021nerf}, to improve performance. Inspired by these methods, we also employ a positionally encoded representation for geometric correspondence during the refinement stage.

\noindent{\textbf{Unseen object pose estimation.}} Some studies focus on category-level pose estimation~\cite{wang2019normalized, chen2020category, manhardt2020cps++, li2022polarmesh}, utilizing shared geometric traits within a category to broaden the range of identifiable target objects. However, these methods face generalization limitations, making it challenging to unseen object categories in real-world settings. To address these limitations, recent research has explored unseen object pose estimation, which aims to accurately predict poses for objects not encountered during training. 

MegaPose~\cite{labbe2022megapose} combines a render-and-compare refiner with a classifier that assesses whether the refiner can correct given pose errors, achieving strong generalization by training on a large synthetic dataset. GigaPose~\cite{nguyen2024gigapose} leverages discriminative templates to handle out-of-plane rotations and uses patch correspondences for estimating remaining pose parameters, resulting in improvements in both speed and segmentation robustness. GenFlow~\cite{moon2024genflow} addresses the accuracy-scalability trade-off by directly leveraging the target object's shape through optical flow prediction. GenFlow iteratively refines poses by leveraging a 3D shape constraint alongside a multi-scale, coarse-to-fine refinement process. FoundPose~\cite{ornek2025foundpose} establishes 2D-3D correspondences by matching patch descriptors from a self-supervised DINOv2~\cite{oquab2023dinov2} model between the image and pre-rendered templates, integrating these descriptors into a bag-of-words model for more efficient template retrieval.

FoundPose is particularly relevant to our approach in that it also aims to estimate correspondences to assist pose estimation. However, FoundPose derives correspondences by performing patch-wise matching with pre-rendered templates, which may result in inaccuracies due to misalignment with the query image. In contrast, our method begins with a coarse pose estimation to establish an initial pose, which we then refine using a rendered reference image closely aligned with the query object. Additionally, rather than relying solely on direct matching, we design a geometry estimation network that improves correspondence estimation by effectively integrating information from the reference image and geometric correspondence guidance.


\section{Method}
\label{sec:method}
This section introduces RefPose, a novel approach to object pose estimation for unseen objects. We start with a brief overview (\cref{subsec:3.1}), followed by a detailed explanation of the initial pose estimation process used to generate a reference (\cref{subsec:3.2}). Finally, we describe how this reference information is leveraged to estimate the query's geometric correspondence and iteratively refine the pose to reach the final result (\cref{subsec:3.3}).

\subsection{Overview}
\label{subsec:3.1}
RefPose follows a multi-stage pipeline, similar to other recent methods~\cite{labbe2022megapose, nguyen2024gigapose, moon2024genflow, ornek2025foundpose}, comprising object detection/segmentation, coarse pose estimation, and pose refinement. Following these methods, we use an off-the-shelf model~\cite{nguyen2023cnos} for object detection and segmentation to preprocess the input image.

\cref{fig:fig2} illustrates the overall pipeline of RefPose. Starting with an RGB image, the target object is detected, cropped, and resized to $256\times256$ to create the query image, $I_{q}$. In the coarse pose estimation stage, a classification network selects a set of templates, $\mathcal{S}=\{S_{1}, S_{2}, \dots, S_{k}\}$, from a pre-rendered template set, $\mathcal{T}=\{T_{1}, T_{2}, \dots, T_{N}\}$. This selected set, $\mathcal{S}$, is then used to estimate the geometric correspondence of the query, $G_{q}$, providing an initial pose estimate, $P_{0}$. Using $P_{0}$, we render a reference image, $I_{r}$, and a geometric correspondence, $G_{r}^{pos}$, that are closely aligned with the query. Here, $G^{pos}$ represents the geometric correspondence with positional encoding applied. In the refinement stage, this reference information is then used to estimate a more accurate geometric correspondence for the query, $G_{q}^{pos}$. Finally, the relative pose, $\Delta P$, between the query and reference is iteratively refined through a render-and-compare approach to reach a final pose estimate.

\begin{figure}[t]
  \centering
   \includegraphics[width=\linewidth]{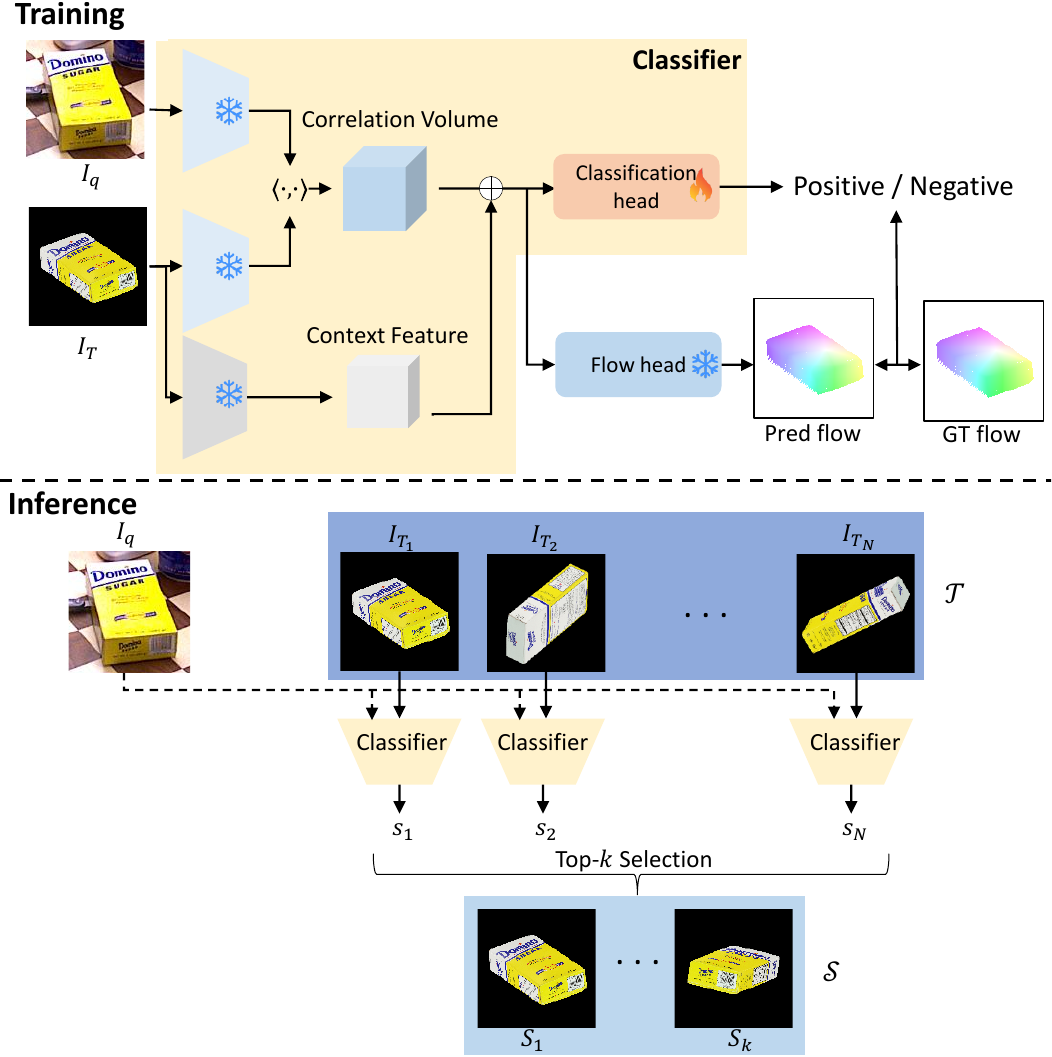}
   \caption{Templates selection using a classification network. The classification network scores pre-rendered templates based on how well optical flow can be estimated between each template and the query image. During inference, these scores are used to select the top-$k$ templates. The classifier leverages a frozen feature encoder from a pre-trained optical flow network, with only the classification head trained.}
   \label{fig:fig3}
\end{figure}

\subsection{Coarse pose estimation}
\label{subsec:3.2}
\noindent{\textbf{Templates selection.}} We start by randomly sampling poses and rendering images along with geometric correspondences for each pose based on the given 3D model. The geometric correspondence $G \in \mathbb{R}^{h \times w \times 3}$ represents a dense map that indicates the corresponding 3D model point for each pixel. For simplicity, we refer to geometric correspondence as "geometry" throughout this paper. Inspired by MegaPose~\cite{labbe2022megapose}, we select a set of multiple templates, $\mathcal{S}$, using a classification network. However, unlike MegaPose, where the classifier is trained based on the refining capability of its refiner, we introduce a new training criterion. Since our subsequent pose estimation stage relies on optical flow, we train the classifier to assess the accuracy of optical flow estimation between each template image, $I_{T}$, and the query image, $I_{q}$.

During training, we utilize a pre-trained optical flow network~\cite{teed2020raft} to estimate the optical flow between $I_{q}$ and each template image in $\mathcal{T}$. Positive and negative pairs are identified by comparing the predicted flow with the ground truth flow, which serves as labels for training the classifier. Additionally, rather than designing and training a new feature encoder for the classification network, we leverage the feature encoder from the optical flow network to enhance both the classifier's performance and coherence with subsequent stages. Specifically, as in RAFT, we extract the correlation volume and context features, then attach a simple CNN as the classification head to complete the classifier. The classifier is trained using Binary Cross-Entropy (BCE) Loss~\cite{ruby2020binary}. During inference, the classifier ranks template images from the set $\mathcal{T}$, and we select the top-$k$ templates based on the predicted scores. The classifier's structure and training and inference processes are illustrated in~\cref{fig:fig3}.

\begin{figure}[t]
  \centering
   \includegraphics[width=\linewidth]{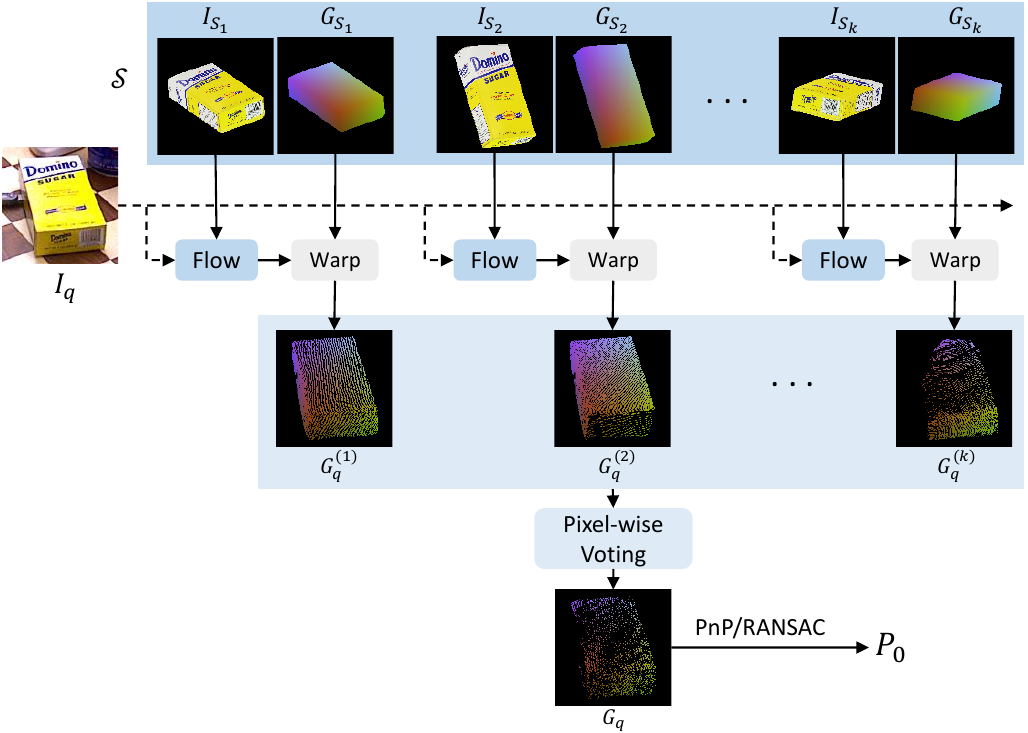}
   \caption{Warping-based geometry estimation process. The optical flow between each template image $I_{S}$ in the selected set $\mathcal{S}$ and the query image $I_{q}$ is used to warp the corresponding template geometries $G_{S}$, generating candidate geometries for the query, $G_{q}$. A pixel-wise voting scheme refines these candidates, and the resulting 2D-3D correspondences in $G_{q}$ are then applied with PnP/RANSAC to estimate the initial pose $P_{0}$.}
   \label{fig:fig4}
\end{figure}

\noindent{\textbf{Warping-based geometry estimation.}} To predict an initial pose $P_{0}$, we estimate the query geometry $G_{q}$ based on a set of selected templates $\mathcal{S}$. First, we calculate the optical flow between each template image $I_{S}$ and the query image $I_{q}$, then warp the template geometries $G_{S}$ accordingly. These warped geometries serve as candidates for $G_{q}$. However, due to potential inconsistencies and inaccuracies in the optical flow estimated from each template, these candidates may indicate different 3D points for the same query pixel. To address this, we employ a voting scheme to select a single 3D point per pixel. PFA~\cite{hu2022perspective} follows a similar approach by aggregating multiple optical flows from different templates to estimate the 2D-3D correspondences. However, it performs this aggregation without explicitly addressing the inconsistencies and inaccuracies, which can lead to unreliable correspondences. In contrast, our method employs a more robust medoid-based voting scheme, where the most representative 3D point per pixel is selected rather than aggregating all candidates indiscriminately. This approach ensures that errors in optical flow estimation do not adversely affect the final result, leading to a more accurate and stable coarse pose estimation. Finally, based on the established 2D-3D correspondences in $G_q$, we apply the PnP/RANSAC algorithm to estimate the pose $P_{0}$. The overall process is illustrated in~\cref{fig:fig4}.

\subsection{Pose refinement}
\label{subsec:3.3}
Using the initial pose $P_{0}$ obtained from the coarse pose estimation stage, we render a single reference image $I_{r}$ and geometry $G_{r}^{pos}$ that are more closely aligned with the query image $I_{q}$ than the previously selected templates in $\mathcal{S}$. Rather than using raw 3D coordinates, we apply positional encoding to the 3D coordinates of $G$ to enrich the geometry representation, thereby improving estimation accuracy~\cite{park2024leveraging}. This encoding results in a representation $G^{pos} \in \mathbb{R}^{h \times w \times 6N_{freq}}$, where $N_{freq}$ denotes the frequency bands used for encoding~\cite{mildenhall2021nerf}.

\noindent{\textbf{Correlation volume-guided attention mechanism.}} Accurately retrieving relevant geometric information from the reference geometry $G_{r}^{pos}$ requires precise pixel-level correspondence between the query image $I_{q}$ and the reference image $I_{r}$. To achieve this, we introduce a correlation volume-guided attention mechanism. In this setup, the query image $I_{q}$ serves as the query, the reference image $I_{r}$ acts as the key, and the reference geometry $G_{r}^{pos}$ is treated as the value. The mechanism computes attention weights based on the similarity between the query ($I_{q}$) and the key ($I_{r}$), enabling the extraction of relevant features from the value ($G_{r}^{pos}$). 

In contrast to vanilla attention mechanisms that implicitly learn relevance, our approach uses explicit pixel-level correspondence information, as both the query image $I_{q}$ and reference image $I_{r}$ represent the same object from different viewpoints. We obtain the attention weights by applying a softmax operation to the correlation volume from the optical flow network, which encodes pixel-level similarities between $I_{q}$ and $I_{r}$. By applying these weights to the reference geometry features $G_{r}^{pos}$, we retrieve highly relevant geometric information, facilitating accurate correspondence estimation for improved pose refinement.

\begin{figure}[t]
  \centering
   \includegraphics[width=\linewidth]{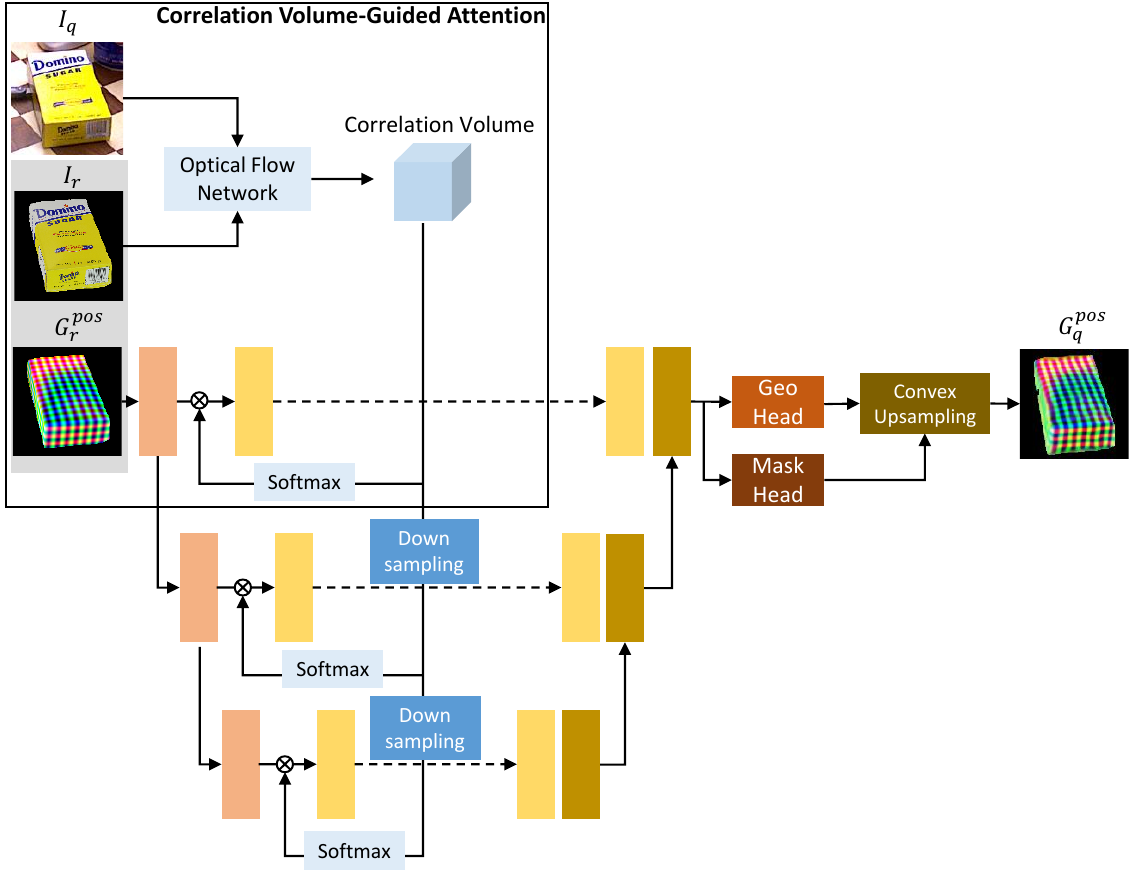}
   \caption{Architecture of the geometry estimation network. The network takes the query image $I_{q}$, reference image $I_{r}$, and reference geometry $G_{r}^{pos}$ as inputs to estimate the query geometry $G_{q}^{pos}$. A correlation volume-guided attention mechanism is applied at each level of the U-Net to effectively integrate these inputs. The geo head and mask head output a low-resolution geometry map and mask, which are then refined through convex upsampling to produce the final high-resolution geometry $G_{q}^{pos}$.}
   \label{fig:fig5}
\end{figure}

\noindent{\textbf{Geometry estimation network.}} Our geometry estimation network processes the query image $I_{q}$, reference image $I_{r}$, and reference geometry $G_{r}^{pos}$ as inputs and outputs the query geometry $G_{q}^{pos}$. To capture multi-scale features effectively, we employ a U-Net structure~\cite{ronneberger2015u}, with correlation volume-guided attention mechanisms applied at each level to incorporate reference information. As features are downsampled within the U-Net, the correlation volume is correspondingly downsampled. The U-Net concludes with two heads: a geo head, which estimates geometry at an 8x downsampled resolution, and a mask head, which predicts an up-mask for convex upsampling. Following RAFT~\cite{teed2020raft}, convex upsampling is applied to reconstruct $G_{q}^{pos}$ at the original resolution, providing higher fidelity than bilinear upsampling. The complete network architecture is illustrated in~\cref{fig:fig5}.

The network is optimized by minimizing the $\mathcal{L}_{1}$ loss, which measures the difference between the predicted and ground truth geometry:
\begin{equation}
\label{eq2}
\mathcal{L}_{geo} = \|\bar{G}_{q}^{pos}-G_{q}^{pos}\|_{1},
\end{equation}
where $\bar{G}_{q}^{pos}$ is the ground truth geometry and $G_{q}^{pos}$ is the predicted geometry. 

\noindent{\textbf{Iterative pose refinement.}} The estimated query geometry $G_{q}^{pos}$ enables iterative pose refinement through a render-and-compare approach~\cite{li2018deepim, labbe2020cosypose, park2022dprost}. Initially, a relative pose $\Delta P$ is estimated between $G_{q}^{pos}$ and the reference geometry $G_{r}^{pos}$, updating the initial pose $P_{0}$. With each update, a new reference geometry is rendered and compared to the fixed query geometry $G_{q}^{pos}$, progressively refining the pose. This iterative process continues until the final pose is accurately determined.

The relative pose estimator, based on a CNN, is trained using sequence loss~\cite{teed2020raft}, which is inspired by SCFlow~\cite{hai2023shape}, to improve learning efficiency and ensure consistent prediction quality across iterations. The sequence loss, $\mathcal{L}_{seq}$, is defined as follows:
\begin{equation}
\label{eq3}
\mathcal{L}_{seq} = \sum\limits_{m=1}^{M}\gamma^{M-m}\mathcal{L}_{pose}^{(m)},
\end{equation}
where $M$ denotes the total number of refinement iterations, $\gamma$ is an exponential weighting factor, and $\mathcal{L}_{pose}^{(m)}$ represents the pose loss at each iteration $m$. 

\renewcommand{\arraystretch}{1.05}
\begin{table*}[!t]
\centering
\caption{Evaluation results on BOP benchmark datasets. The table reports Average Recall (AR) scores across the seven datasets in the BOP challenge, where higher AR scores indicate better performance. The best-performing method is highlighted in bold, and the second-best is underlined. The top section presents results from coarse pose estimation alone, while the bottom section displays results after applying the refinement stage. ``MH'' denotes MegaPose and GenFlow versions that incorporate a multi-hypotheses strategy in the refinement stage, and ``featuremetric'' refers to the refinement method introduced in FoundPose.} 
{\resizebox{0.9\linewidth}{!}
{\begin{tabular}{c|c|ccccccc|c|c}
\hline
\textbf{Method} & \textbf{Refinement} & \textbf{YCB-V} & \textbf{LM-O} & \textbf{T-LESS} & \textbf{TUD-L} & \textbf{IC-BIN} & \textbf{ITODD} & \textbf{HB} & \textbf{Mean} & \textbf{Run-time} \\
\hline
OSOP~\cite{shugurov2022osop} & - & 29.6 & 27.4 & \textbf{40.3} & - & - & - & -  & - & - \\
ZS6D~\cite{ausserlechner2024zs6d}  & - & 32.4 & 29.8 & 21.0 & - & - & - & -  & - & - \\
MegaPose~\cite{labbe2022megapose} & - & 28.1 & 22.9 & 17.7 & 25.8 & 15.2 & 10.8  & 25.1 & 20.8 & 15.5s \\
GenFlow~\cite{moon2024genflow}  & - & 27.7 & 25.0 & 21.5 & 30.0 & 16.8 & 15.4 & 28.3 & 23. & 3.8s \\
GigaPose~\cite{nguyen2024gigapose}  & - & 27.8 & 29.6 & 26.4 & 30.0 & 22.3 & 17.5 & 34.1 & 26.8 & \textbf{0.4s} \\
FoundPose~\cite{ornek2025foundpose}  & - & \underline{45.2} & \textbf{39.6} & 33.8 & \underline{46.7} & \textbf{23.9} & \underline{20.4} & \textbf{50.8} & \underline{37.2} & \underline{1.7s} \\
RefPose (Ours)  & - & \textbf{50.0} & \underline{35.8} & \underline{38.1} & \textbf{48.5} & \underline{23.1} & \textbf{21.5} & \underline{49.8} & \textbf{38.1} & 3.1s \\
\hline
MegaPose ~\cite{labbe2022megapose}  & MegaPose & 60.1 & 49.9 & 47.7 & 65.3 & 36.7 & 31.5 & 65.4  & 50.9 & 17.0s \\
MegaPose ~\cite{labbe2022megapose}  & MegaPose, MH & 62.1 & 56.0 & 50.7 & 68.4 & 41.4 & 33.8 & 70.4  & 54.7 & 21.9s \\
MegaPose ~\cite{labbe2022megapose}  & RefPose (Ours) & 65.3 & 56.0 & 52.8 & 66.4 & 45.3 & 41.1 & 73.6 & 57.2 & 16.4s \\
GenFlow ~\cite{moon2024genflow}  & GenFlow, MH & 63.3 & 56.3 & 52.3 & 68.4 & 45.3 & 39.5 & 73.9 & 57.0 & 20.8s \\
GigaPose ~\cite{nguyen2024gigapose}  & MegaPose & 63.2 & 55.7 & 54.1 & 58.0 & 45.0 & 37.6 & 69.3  & 54.7 & \textbf{2.3s} \\
GigaPose ~\cite{nguyen2024gigapose}  & GenFlow, MH & 65.2 & \textbf{63.1} & \textbf{58.2} & 66.4 & \underline{49.8} & \textbf{45.3} & \underline{75.6}  & \underline{60.5} & 10.6s \\
FoundPose ~\cite{ornek2025foundpose}  & MegaPose, MH + Featuremetric & \underline{69.0} & \underline{61.0} & 57.0 & \underline{69.4} & 47.9 & 40.7 & 72.3 & 59.6 & 20.5s \\
RefPose (Ours) & MegaPose & 63.7 & 56.3 & 51.1 & 65.8 & 43.7 & 41.4 & 71.8 & 56.3 & 4.6s \\
RefPose (Ours)  & RefPose (Ours) & \textbf{72.7} & 59.6 & \underline{57.8} & \textbf{69.7} & \textbf{51.2} & \underline{43.8} & \textbf{76.2} & \textbf{61.4} & \underline{3.9s} \\
\hline
\end{tabular}
}

\label{table:tab1}
}
\end{table*}

After updating the reference pose, we compare it with the ground truth pose to assess the refinement strategy. The pose loss $\mathcal{L}_{pose}$, which measures this alignment, combines grid-matching and grid-distance loss functions~\cite{park2022dprost}, formulated as follows:
\begin{equation}
\label{eq4}
\mathcal{L}_{pose} = \|\bar{\mathcal{G}}-\mathcal{G}\|_{2} + \|\|\bar{t}\|_{2} - \|t\|_{2}\|_{1},
\end{equation}
where $ \bar{\mathcal{G}} $ and $ \bar{t} $ are the ground truth grid and translation vectors, respectively, while $\mathcal{G}$ and $t$ are derived from the estimated pose.

\section{Experiment}
\label{sec:experiment}
\subsection{Experimental Setup}
\label{subsec:4.1}
\noindent{\textbf{Datasets.}} We train our model on the synthetic dataset, Google Scanned Objects (GSO)~\cite{downs2022google}, as provided by~\cite{labbe2022megapose}. This dataset contains nearly 1 million images representing 1,000 different object types. Although~\cite{labbe2022megapose} also includes data from approximately 50,000 ShapeNet~\cite{chang2015shapenet} objects, we opted to use only the GSO dataset due to computational and memory constraints. This choice is further supported by the higher-quality mesh models in the GSO dataset, which, as noted in~\cite{labbe2022megapose}, contribute more critically to model performance than ShapeNet's. For evaluation, we test our approach on the seven primary datasets in the BOP benchmark~\cite{hodan2024bop}, including YCB-V, LM-O, T-LESS, TUD-L, ICBIN, ITODD, and HB. Our method relies solely on RGB images and 3D object models, without leveraging any depth information.

\noindent{\textbf{Evaluation metrics.}} We follow the standard BOP evaluation protocol~\cite{hodan2024bop}, which employs three core metrics: Visible Surface Discrepancy (VSD), Maximum Symmetry-Aware Surface Distance (MSSD), and Maximum Symmetry-Aware Projection Distance (MSPD). Overall performance, referred to as Average Recall (AR), is calculated by averaging the individual recall scores for each metric across a range of error thresholds.

\noindent{\textbf{Implementation details.}} Our model is trained with the AdamW optimizer~\cite{loshchilov2017decoupled}, using a batch size of 8 and a learning rate of 0.0001 for a total of 400k training steps. A cosine annealing scheduler~\cite{loshchilov2016sgdr} with a 10k step period is employed to adjust the learning rate. Both training and evaluation are conducted on an RTX-3090 GPU. For the optical flow network in RefPose, we use the large model of RAFT~\cite{teed2020raft}, which is pre-trained on the FlyingChairs~\cite{dosovitskiy2015flownet} and FlyingThings3D~\cite{mayer2016large} datasets and then fine-tuned on the GSO dataset for our specific application. In the pose refinement stage, we generate the geometry $G^{pos}$ using sine and cosine positional encoding with five frequency bands, following the approach in~\cite{park2024leveraging}. The number of pose refinement iterations, $M$, is set to 5, balancing accuracy and runtime efficiency.

\captionsetup[subfigure]{labelformat=empty}
\begin{figure}[t]
  \centering
    \begin{subfigure}{0.17\linewidth}
        \includegraphics[width=1.0\linewidth]{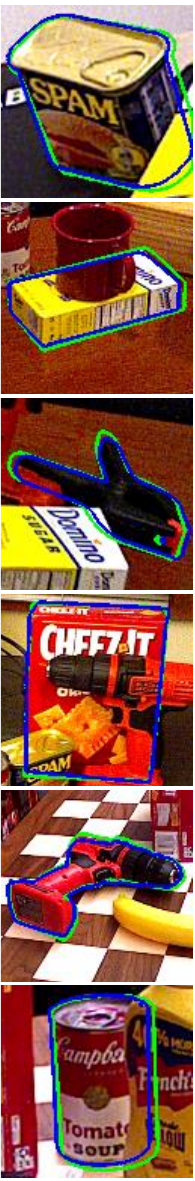}
        \subcaption{Ours}
    \end{subfigure}
    \begin{subfigure}{0.17\linewidth}
        \includegraphics[width=1.0\linewidth]{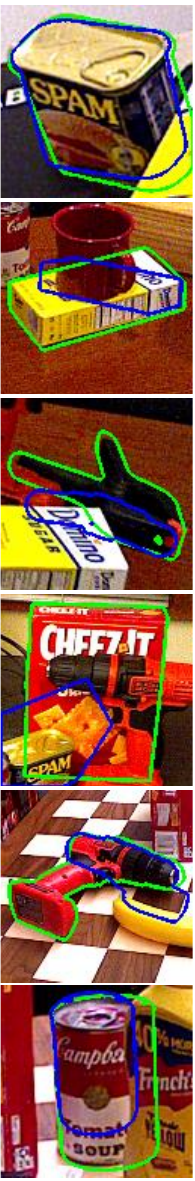}
        \subcaption{FoundPose}
    \end{subfigure}
    \begin{subfigure}{0.17\linewidth}
        \includegraphics[width=1.0\linewidth]{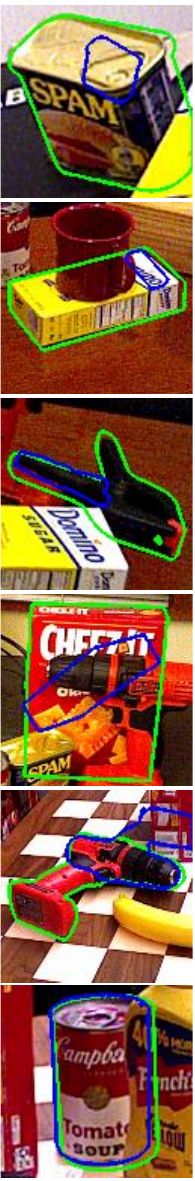}
        \subcaption{GigaPose}
    \end{subfigure}
    \begin{subfigure}{0.17\linewidth}
        \includegraphics[width=1.0\linewidth]{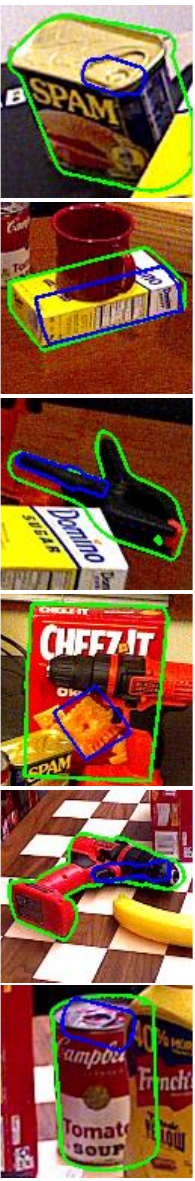}
        \subcaption{GenFlow}
    \end{subfigure}
    \begin{subfigure}{0.17\linewidth}
        \includegraphics[width=1.0\linewidth]{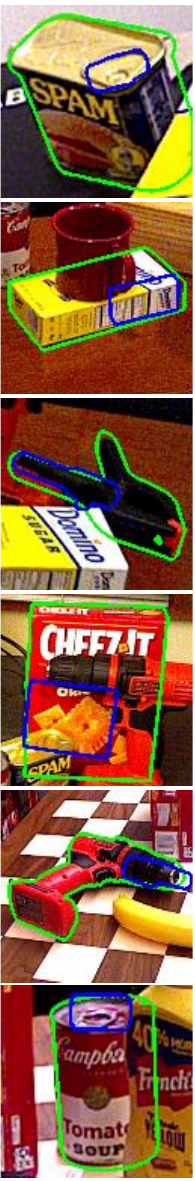}
        \subcaption{MegaPose}
    \end{subfigure}
    \caption{Qualitative comparison of pose estimation results. We present a qualitative comparison of our method against other approaches, with the projected contours from the ground-truth pose shown in green and those from the predicted pose in blue.}  
   \label{fig:fig6}
\end{figure}

\subsection{Comparison with state-of-the-art methods}
\label{subsec:4.2}
\cref{table:tab1} presents the results of our method on the BOP benchmark datasets. Except for OSOP~\cite{shugurov2022osop}, which uses its own detection model, all methods employ CNOS~\cite{nguyen2023cnos} as the detection/segmentation model to ensure a fair comparison. Our method achieves the best performance in both the coarse pose estimation and the refined results following the refinement stage. While our approach slightly underperforms the current state-of-the-art methods on LM-O, T-LESS, and ITODD, it achieves state-of-the-art results on the other datasets, with particularly notable improvements on YCB-V and HB. Overall, our method demonstrates the best performance across all datasets.~\cref{fig:fig6} provides qualitative comparisons with existing methods, further illustrating the robustness of our approach.

The 10th and 15th rows of~\cref{table:tab1} report results where our proposed coarse pose estimation and refinement methods are independently combined with MegaPose to assess their standalone effectiveness. A comparison among the 8th, 12th, and 15th rows shows that our coarse pose estimation achieves better results even when using the same refinement method as MegaPose. Additionally, comparing the 8th and 10th rows demonstrates that our refinement method is more effective even when applied to the coarse poses estimated by MegaPose. 

The reported runtime represents the average speed for processing all objects within a single image. Though our method takes slightly longer in the coarse pose estimation stage, it significantly reduces refinement time by minimizing the number of renderings and using a lightweight model. As a result, our method achieves superior performance with runtime comparable to other state-of-the-art approaches.

\subsection{Ablation study}
\label{subsec:4.3}
\noindent{\textbf{Number of pre-rendered templates.}} The results of the ablation study on the number of pre-rendered templates in the set $\mathcal{T}$, represented by $N$, are shown in~\cref{table:tab2}. With only 64 templates, the sampled poses are too sparse across the object's orientation space, leading to reduced performance in pose estimation due to insufficient coverage of various possible object poses. Increasing to 128 templates provides a denser sampling, significantly enhancing query-template matching and improving performance. Using more than 128 templates yields only marginal improvements while increasing both memory and computational overhead. Additionally, a larger template set results in longer inference times, as more templates must be evaluated. Therefore, using 128 templates achieves an optimal balance between efficiency and performance.

\begin{table}[t]
\centering
    \centering
    \caption{Ablation study on the number of pre-rendered templates. The table presents AR scores, illustrating the impact of varying numbers of pre-rendered templates, $\mathcal{T}$, on coarse pose estimates.}
    {\resizebox{0.95\linewidth}{!}
    {\begin{tabular}{c|ccccccc|c}
        \hline
        $N$ & YCB-V  & LM-O & T-LESS & TUD-L & IC-BIN & ITODD & HB & Mean                                      \\
        \hline
        64 & 45.2 & 29.8 & 36.5 & 43.7 & 20.1 & 18.8 & 43.5 & 33.9 \\
        128 & \textbf{50.0} & 35.8 & 38.1 & \textbf{48.5} & 23.1 & 21.5 & \textbf{49.8} & 38.1 \\
        256 & 48.8 & \textbf{37.2} & \textbf{39.5} & 45.8 & \textbf{24.3} & \textbf{23.1} & \textbf{49.8} & \textbf{38.3} \\
        \hline
    \end{tabular}}
     }
    \label{table:tab2}
\end{table}

\begin{table}[t]
\centering
    \centering
    \caption{Ablation study on the number of selected templates. The table presents AR scores, showing the impact of varying numbers of selected templates, $\mathcal{S}$, on coarse pose estimates.}
    {\resizebox{0.95\linewidth}{!}
    {\begin{tabular}{c|ccccccc|c}
        \hline
        $k$ & YCB-V  & LM-O & T-LESS & TUD-L & IC-BIN & ITODD & HB & Mean                                      \\
        \hline
        1 & 42.6 & 31.0 & 28.8 & 40.1 & 18.8 & 18.1 & 40.8 & 31.5  \\
        2 & 46.8 & \textbf{37.2} & 34.0 & 43.1 & 19.8 & 18.8 & 46.5 & 35.2 \\
        4 & \textbf{50.0} & 35.8 & \textbf{38.1} & \textbf{48.5} & \textbf{23.1} & \textbf{21.5} & 49.8 & \textbf{38.1} \\
        8 & \textbf{50.0} & 29.8 & 36.4 & 44.5 & 21.7 & 19.9 & \textbf{51.0} & 36.2 \\
        \hline
    \end{tabular}}
    \label{table:tab3}}
\end{table}

\noindent{\textbf{Number of selected templates.}} The results of our ablation study on the number of selected templates $\mathcal{S}$, denoted by $k$, are shown in~\cref{table:tab3}. Selecting four templates from the pre-rendered set $\mathcal{T}$ yields the best results, achieving a balance between diversity and alignment accuracy. Using fewer templates limits diversity, increasing the risk of alignment errors during warping-based geometry estimation if a template is poorly selected or if the optical flow is inaccurately estimated potentially, leading to inaccurate pose estimation. Conversely, selecting more than four templates raises the likelihood of including misaligned templates, which may reduce the effectiveness of the medoid-based voting stage in handling outliers. With too many templates, the medoid's robustness can be compromised, as it becomes more challenging to filter out misaligned correspondence effectively. Thus, selecting four templates provides an effective balance, ensuring sufficient diversity while minimizing alignment errors, which is crucial for accurate geometry and pose estimation.

\noindent{\textbf{Components in the coarse pose estimation stage.}}~\cref{table:tab4} presents the results of our ablation study on key components used in the coarse pose estimation stage. The first row evaluates the impact of using the feature encoder from the optical flow network as the classifier's feature encoder. This feature encoder outputs flow features, including the correlation volume and context feature, which provide rich cues related to optical flow, enhancing the classifier's accuracy in selecting templates. By directly leveraging the same encoder applied in the warping-based geometry estimation stage, consistency across stages is maintained, contributing to superior results.

\begin{table}[t]
\centering
    \centering
    \caption{Ablation study on components in coarse pose estimation stage. This table presents AR scores for the coarse pose estimates.}
    {\resizebox{0.95\linewidth}{!}
    {\begin{tabular}{c|ccccccc|c}
        \hline
        Setting & YCB-V  & LM-O & T-LESS & TUD-L & IC-BIN & ITODD & HB & Mean                                      \\
        \hline
        w/o Flow features & 46.8 & 34.2 & 33.2 & 47.1 & 22.3 & 20.9 & 45.8 & 35.8 \\
        w/o Medoid & 49.6 & 29.6 & 36.8 & 45.9 & 21.3 & 18.8 & \textbf{50.2} & 36.0 \\
        Ours & \textbf{50.0} & \textbf{35.8} & \textbf{38.1} & \textbf{48.5} & \textbf{23.1} & \textbf{21.5} & 49.8 & \textbf{38.1} \\
        \hline
    \end{tabular}}
    \label{table:tab4}}
\end{table}

\begin{table}[t]
\centering
    \centering
    \caption{Ablation study on components of the geometry estimation network in the pose refinement stage. This table presents AR scores for the pose refinement results.}
    {\resizebox{0.95\linewidth}{!}
    {\begin{tabular}{c|ccccccc|c}
        \hline
        Setting & YCB-V  & LM-O & T-LESS & TUD-L & IC-BIN & ITODD & HB & Mean                                      \\
        \hline
        w/o P.E.  & 69.3 & 57.1 & 57.0 & 68.4 & 48.3 & 39.5 & 73.9 & 59.1 \\
        w/o Convex. & 70.0  & 56.8 & \textbf{58.2} & 68.4 & 47.9 & 43.1 & 75.6 & 60.0 \\
        w/o C.G. attn.& 68.8 & \textbf{61.0} & 52.8 & 66.0 & 47.3 & 41.9 & 74.8 & 58.9 \\
        Ours& \textbf{72.7} & 59.6 & 57.8 & \textbf{69.7} & \textbf{51.2} & \textbf{43.8} & 
        \textbf{76.2} & \textbf{61.4} \\
        \hline
    \end{tabular}}
     
    \label{table:tab5}}
\end{table}

The ``w/o Medoid'' variant represents a model in which, during pixel-wise voting in the warping-based geometry estimation, the medoid is replaced with a simple average for selecting correspondences. Using the medoid rather than averaging mitigates the impact of outliers that may arise from imperfect optical flow estimations, leading to a more robust correspondence selection. This robustness directly translates to greater accuracy of the coarse pose estimation.

\noindent{\textbf{Components of the geometry estimation network in the pose refinement stage.}}~\cref{table:tab5} presents the ablation study results on key components of the geometry estimation network within the pose refinement stage. The first row shows that applying positional encoding to the 3D coordinates, inspired by~\cite{park2024leveraging}, improves performance over using raw 3D coordinates as geometric correspondence. Specifically, the ``w/o P.E.'' variant, which omits positional encoding and directly uses 3D coordinates, shows lower accuracy. 

Additionally, estimating an up-mask and utilizing convex upsampling yield superior performance compared to bilinear upsampling, as indicated by the ``w/o Convex.'' variant. Convex upsampling more effectively preserves spatial detail at the original resolution, leading to a closer alignment between the estimated geometry and the true geometry of the object.

\begin{figure}[t]
  \centering
    \begin{subfigure}{0.22\linewidth}
        \includegraphics[width=1.0\linewidth]{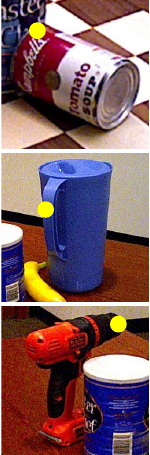}
        \subcaption{$I_q$}
    \end{subfigure}
    \begin{subfigure}{0.22\linewidth}
        \includegraphics[width=1.0\linewidth]{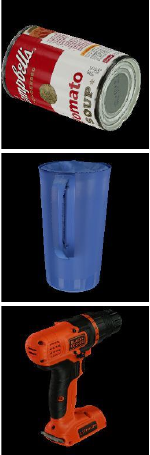}
        \subcaption{$I_r$}
    \end{subfigure}
    \begin{subfigure}{0.22\linewidth}
        \includegraphics[width=1.0\linewidth]{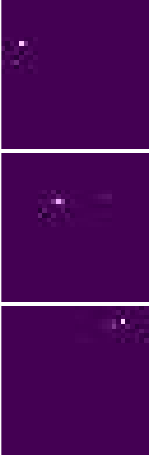}
        \subcaption{C.G. Attn.}
    \end{subfigure}
    \begin{subfigure}{0.22\linewidth}
        \includegraphics[width=1.0\linewidth]{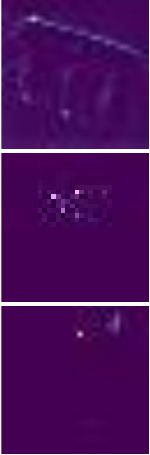}
        \subcaption{Vanilla Attn.}
    \end{subfigure}
    \caption{Visual comparison of attention mechanisms in the geometry estimation network. Each row shows a query image $I_{q}$ with a yellow dot marking the point of interest, the corresponding reference image $I_{r}$, and the attention maps produced by the correlation volume-guided (C.G.) attention and vanilla attention. The C.G. attention accurately focuses on relevant regions in $I_{q}$ and $I_{r}$, while vanilla attention lacks this precision.}  
   \label{fig:fig7}
\end{figure}

Finally, the proposed correlation volume-guided (C.G.) attention mechanism outperforms vanilla attention, as demonstrated by the ``w/o C.G. attn.'' variant, which replaces C.G.attention with vanilla attention. This result underscores the effectiveness of C.G. attention in accurately capturing relevant correspondences between the query and reference images. As illustrated in~\cref{fig:fig7}, C.G. attention maps exhibit a sharper focus on relevant regions in the reference image that correspond to points of interest in the query, while vanilla attention maps appear less precise. This comparison highlights the effectiveness of C.G. attention in establishing accurate correlations between the query and reference images.

\captionsetup[subfigure]{labelformat=empty}

\section{Conclusion}
\label{sec:conclusion}
In this paper, we have proposed RefPose, a two-stage method designed for enhanced accuracy and generalization in unseen object pose estimation. Starting with a coarse pose estimation using template selection and medoid-based voting, RefPose builds an initial pose, which is then used to assist the geometry estimation through a correlation volume-guided attention mechanism. This refined geometry for the query supports an iterative render-and-compare process, producing a precise final pose. Extensive experiments on the BOP benchmark demonstrate RefPose's strong performance and generalization ability, while ablation studies confirm the effectiveness of each component. RefPose advances adaptable and efficient solutions for 6D pose estimation in complex, real-world scenarios.

\noindent\textbf{Acknowledgements} This work was supported by Samsung Electronics Co., Ltd.(No. 0423-20240056), Institute of Information \& communications Technology Planning \& Evaluation (IITP) grant funded by the Korea government(MSIT) [NO.RS-2021-II211343, Artificial Intelligence Graduate School Program (Seoul National University)], and in part by the BK21 FOUR program of the Education and Research Program for Future ICT Pioneers, Seoul National University in 2025.


{
    \small
    \bibliographystyle{ieeenat_fullname}
    \bibliography{pose}
}


\end{document}